\pdfoutput=1

\documentclass[11pt]{article}

\usepackage[final]{acl}

\usepackage{times}
\usepackage{latexsym}

\usepackage[T1]{fontenc}

\usepackage[utf8]{inputenc}

\usepackage{microtype}

\usepackage{inconsolata}
\usepackage{amsmath}

\usepackage{xurl}
\usepackage{booktabs}
\usepackage{multirow}

\usepackage{tikz}
\usepackage{pgfplots}
\usepackage{pgf-pie}
\usepackage{subcaption}
\usepackage{soul}

\pgfplotsset{compat=1.18} 
\usetikzlibrary{patterns}
\usetikzlibrary{positioning}
\definecolor{bar1}{RGB}{68,114,196}
\definecolor{bar2}{RGB}{187,100,146}
\definecolor{bar3}{RGB}{165,165,165}
\definecolor{bar4}{RGB}{255,192,0}
\definecolor{bar5}{RGB}{91,155,213}
 \newcounter{i}
 \def\mycount{\Alph{i}\ifnum\value{i}<6
  \stepcounter{i}
  \else
  \setcounter{i}{1}
  \fi\kern-1ex}

\newcommand{\cameraready}[1]{#1}

\title{Explainability and Hate Speech:\\Structured Explanations Make Social Media Moderators Faster}

\author{Agostina Calabrese$^{1}$\thanks{This work was done while the author was an intern at Snap Inc.} Leonardo Neves$^{2}$ Neil Shah$^{2}$ Maarten W. Bos$^{2}$ Bj{\"o}rn Ross$^{1}$ \\ \textbf{Mirella Lapata}$^{1}$ \textbf{Francesco Barbieri}$^{2}$ \\
        School of Informatics, University of Edinburgh$^{1}$  Snap Inc.$^{2}$ \\ a.calabrese@ed.ac.uk}

\begin{document}
\maketitle
\begin{abstract}
Content moderators play a key role in keeping the conversation on social media healthy. While the high volume of content they need to judge represents a bottleneck to the moderation pipeline, no studies have explored how models could support them to make faster decisions.
There is, by now, a vast body of research into detecting hate speech, sometimes explicitly motivated by a desire to help improve content moderation, but published research using real content moderators is scarce.
In this work we investigate the effect of explanations on the speed of real-world moderators. Our experiments show that while generic explanations do not affect their speed and are often ignored, structured explanations 
lower moderators' decision making time by 7.4\%.
\end{abstract}

\section{Introduction}
Social media provide a platform for free expression but users may abuse it and post content in violation of terms, like misinformation or hate speech. To fight these behaviours and enforce integrity on the platform, social media companies define policies that describe what content is allowed. Posts are then monitored through automatic systems that look for policy violations. While content that has been flagged by the system with high confidence is immediately removed, all other violations, including the ones reported by users, are \emph{moderated} by trained \textit{human} reviewers. These moderators are also responsible for reviewing user appeals and deciding when content has been flagged incorrectly. Therefore, a big challenge with enforcing integrity is the high volume of content that needs to pass the moderators' judgment \cite{2022-Halevy-social}.

Previous work has claimed that moderators can be supported with explanations of why posts violate the policy \citep{2022-calabrese-plead,2023-nguyen-fairexpl}. But while there have been studies showing the importance of explanations for users \cite{2021-haimson-disproportionate,2019-Brunk-trust}, the benefits of explanations for moderators have not been studied. Can explanations help moderators judge a post faster?
And how much room for improvement is there?
While social media share safety reports with statistics about the number and types of detected violations\footnote{e.g., \url{https://about.fb.com/news/2023/05/metas-q1-2023-security-reports}}, data relative to moderator performance is not publicly available.
Explanations might have a larger impact on the performance of crowdworkers who have only recently been trained on a policy, but smaller effects would be expected on the speed of moderators who know the policy by heart.
%

In this paper we conduct a study with \emph{professional} moderators from an online social platform to answer the following research questions:
\begin{enumerate}
    \item Do explanations make moderators faster?
    \item Does the type of explanations matter?
    \item Do moderators want explanations?
\end{enumerate}
While online social platforms deal with several integrity issues, academic research has focused on a few specific ones. Hate speech is one of the most studied issues, and (English) hate speech is also the focus of our study.
Our experiments show that despite their already impressive performance, structured explanations (that highlight which parts of a post are harmful and why) can make \emph{experienced} moderators faster by 1.34s/post without any loss in accuracy. Considering that they spend an average of 18.14s/post, that is a time reduction of 7.4\%, which is a meaningful improvement considering the scale at which online social platforms operate. Generic (pre-defined) explanations on the other hand have no impact.

An online survey further revealed that moderators strongly prefer 
structured explanations (84\%). 
In the case of generic explanations, most moderators admit to only looking at them when in doubt (80\%) or ignoring them completely (12\%). 

\section{Related Work}
While some researchers have looked at hate speech\footnote{``Abusive speech targeting specific group characteristics, such as ethnic origin, religion, gender or sexual orientation'' \cite{2012-warner-hatespeech}.} as a subjective matter  \cite{2022-davani-disagreement,2021-basile-disagreement}, this paradigm is not suitable for the use case of content moderation, where a single decision has to be made for each post \cite{2022-rottger-paradigm}. In this work we follow a prescriptive paradigm, and assume the existence of a ground truth that is determined by a policy.

Explainability is a key open problem for Natural Language Processing research on hate speech \cite{2019-mishra-survey,2021-mathew-hatexplain}. Well documented model failures \cite{2019-sap-bias,2021-calabrese-aaa}, together with EU regulations on algorithmic transparency \cite{2019-Brunk-trust}, call for the design of more transparent algorithms. 
However, the benefits of explainability on the moderators have been understudied. \citet{2023-wang-explanations} analysed the effect of explanations on annotators, observing that wrong explanations might dangerously convince the annotators to change their mind about whether a post contains hate speech. However, the experiment was run with crowdworkers and \citet{2023-abercrombie-temporal} has found that is not uncommon for non-professional moderators to change their opinion about the toxicity of a post over time, even when no additional information is provided. To the best of our knowledge, we are the first to explore how explanations can affect moderation speed 
of professional moderators although the need to support them with their unmanageable workload is well-documented\footnote{e.g., \url{https://www.forbes.com/sites/johnkoetsier/2020/06/09/300000-facebook-content-moderation-mistakes-daily-report-says/?sh=524ab91354d0} and \url{https://www.wired.co.uk/article/facebook-content-moderators-ireland}}.

\section{Explainable Abuse Detection}
We hypothesise that different types of explanations might lead to different results. \citet{2019-mishra-survey} argue that explanations should at least indicate 1) the intent of the user, 2) the words that constitute abuse, and 3) who is the target. From a computational perspective, the cheapest way to achieve this goal is to define the task as multiple multi-class classification problems \cite{2023-kirk-semeval,2021-saeidi-tree,2021-vidgen-lftw,2019-ousidhoum-mlma}, where models choose between some predefined target groups (e.g., women, lgbt+) and types of abuse (e.g., threats, derogation).
While the explanations provided by these approaches are limited to properties 1 and 3, some approaches have expanded the paradigm to also include rationales (i.e., spans of text from the post that suggest why a post is hateful) and satisfy property 2 \cite{2021-vidgen-cad,2021-mathew-hatexplain}.
When dealing with implicit hate, where evidence cannot always be found in the exact words of a post, rationales have been replaced with free-text implied statements \cite{2021-elsherief-latent,2020-sap-sbic}.
\citet{2022-calabrese-plead} introduce a more structured approach to explainability, where target, intent, and type of abuse are all indicated by means of \textit{tagged} spans from the post. 
The popularity of prompt-based approaches has led to the generation of free-text explanations \cite{2023-wang-explanations}, with no guarantee that any of the above properties are satisfied. 

\section{Experimental Design}
In this study we analyse the effect explanations have on the speed of professional moderators from an online social platform with millions of users.
\cameraready{We use the term ``generic''  to describe explanations that can be obtained from a multi-class classification model. For instance, for the post \textit{``immigrants are parasites''}\footnote{Example taken from \citet{2022-calabrese-plead}.}, a generic explanation could be \textit{``Content targeting a person or group of people on the basis of their protected characteristic(s) with dehumanising speech in the form of comparisons, generalisations or unqualified behavioural statements to or about insects''}\footnote{\url{https://transparency.fb.com/en-gb/policies/community-standards/hate-speech}}. This pre-defined explanation illustrates why the post violates the policy without reference to specific post content.
``Structured'' explanations are instead specific to the post, and indicate why a post violates the policy by highlighting  relevant spans and specifying how they relate to the policy. In the framework introduced in \citet{2022-calabrese-plead}, the example above would be associated with a parse tree where \textit{``immigrants''} is tagged as target and protected characteristic, and \textit{``are parasites''} as dehumanising comparison.}
\begin{figure*}[h]
    \centering
    \includegraphics[width=\textwidth]{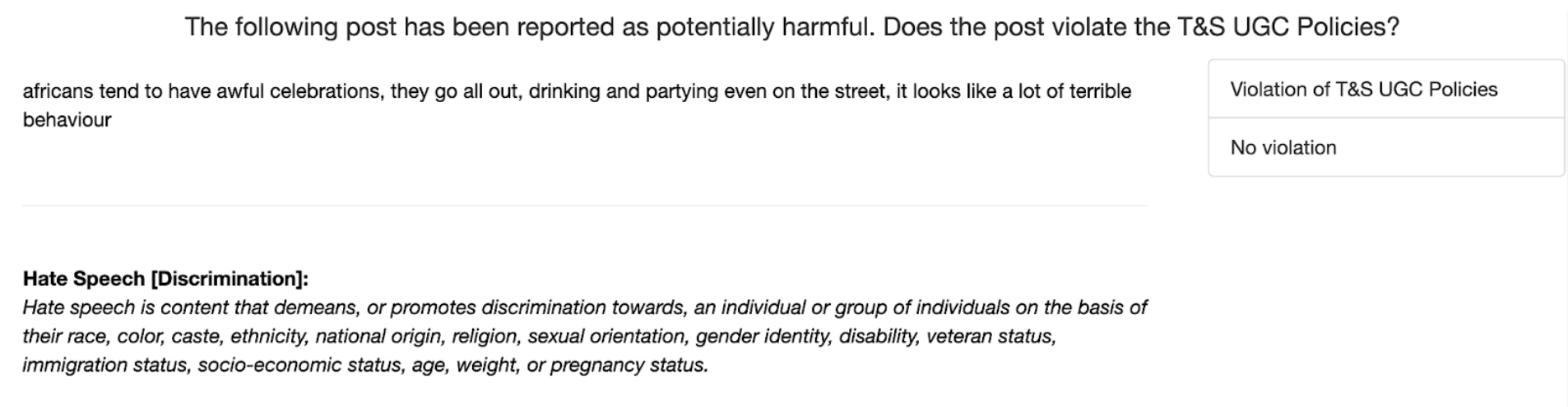}
    \caption{Annotation interface for setting 2 (post+label), where moderators are shown a post and a description of the rule it is deemed to violate. We intentionally chose a generic policy paragraph for this example as we are not allowed to share the content of the internal policies.}
    \label{fig:s2}
\end{figure*}
\begin{figure*}[h]
    \centering
    \includegraphics[width=\textwidth]{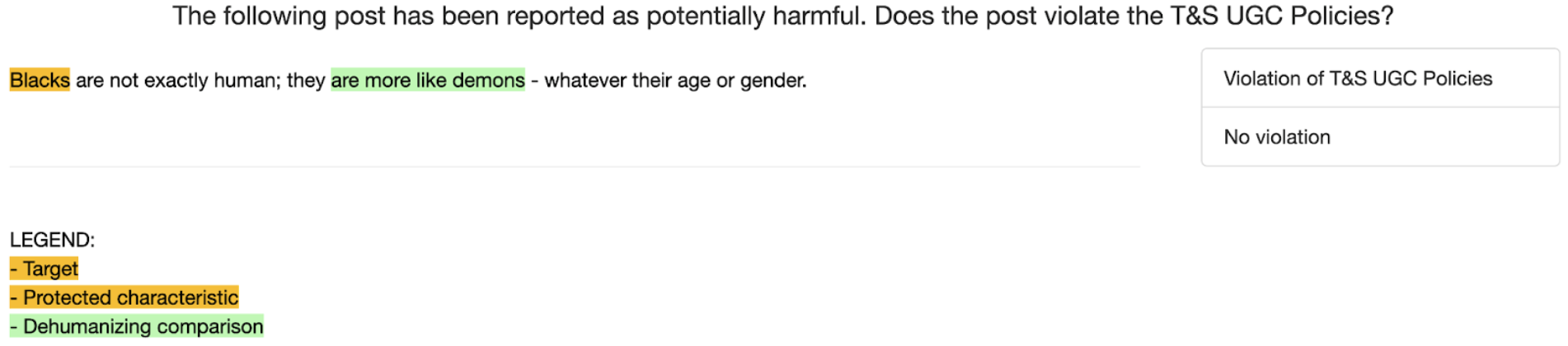}
    \caption{Annotation interface for setting 3 (post+tags), where moderators are shown the post with tagged spans as in \citet{2022-calabrese-plead}.}
    \label{fig:s3}
\end{figure*}
Our hypothesis is that structured explanations will help moderators judge posts faster, while generic explanations will not impact their speed. To verify our hypothesis, we asked 25 moderators to judge posts in three settings where they were shown: 1) only the post (\textbf{post-only}); 2) the post and the policy rule being violated (\textbf{post+policy}, which we refer to as generic explanations, \cameraready{Figure \ref{fig:s2}}) \cite{2023-kirk-semeval}; 3) the post with tagged spans as in \citet{2022-calabrese-plead} (\textbf{post+tags}, that is, structured explanations, \cameraready{Figure \ref{fig:s3}}).

\subsection{Data}
For our experiment we used the PLEAD dataset \cite{2022-calabrese-plead}.
PLEAD contains 3,535 hateful and not-hateful posts annotated with the user intent (e.g., dehumanisation) and explanations in the form of parse trees. 
We include more details about PLEAD in Appendix~\ref{sec:plead}.

\cameraready{While there exist models that can generate structured explanations, the best model available in the literature achieved a production F1-score of 52.96\% \cite{2022-calabrese-plead}. We argue that using generated explanations in our study would bias the results. If the model gives wrong explanations half the time, then that prevents us from measuring how useful correct explanations are, or what ``type'' of explanations is most useful. In light of this, we used gold explanations from the PLEAD dataset.}

Since moderators would normally check posts that are ``at risk'', we reproduced their usual task by mostly sampling hateful posts. However, to keep the experiment realistic, we simulated some model errors: in each of the three settings we included posts that do not violate the policy (10\%); posts that violate the policy but are shown together with wrong explanations (10\%); the remaining posts are hateful (80\%) and associated with the explanations from the dataset. 
While the simulated model accuracy is high, with 80\% correct explanations and 90\% correct predictions, we feared that trivial errors would still push the moderators towards ignoring the explanations \cite{2015-dietvorst-err}. To mitigate this issue, we first used heuristics to generate better explanations 
and then manually reviewed and edited the modified explanations (Appendix \ref{sec:error-simulation}).
We sampled a batch of 100 posts for a pilot study and three batches of 800 posts for the final experiment, one for each setting. The distribution of the intents 
in each setting is the same as in PLEAD.

\subsection{Method}
\cameraready{We recruited 25 moderators from Snapchat, an online social platform with millions of users. All moderators had experience reviewing posts with  abusive language (as the platform policies are wider and contain many more phenomena) and posts that only contain text (as most moderators at the platform usually deal with multimodal content). We recognise that different levels of moderators experience might lead to different results. None of our moderators were new hires. Furthermore, we used mixed-effects models to analyse our results as a way to take into account different levels of experience and therefore ``baseline'' speed.}

We asked moderators to annotate 2,400 posts, 800 for each setting, thus preventing moderators from encountering the same post twice and bias speed measurements. 
The order in which the settings were shown to moderators was randomised. Some moderators received setting 1 first, others received setting 2 first, etc. Each setting was shown as the first setting roughly the same number of times (respectively 8, 8 and 9). Each block of 800 posts was used for each setting a third of the time. \cameraready{This means that the observed results do not depend on the specific posts that occur in a block, because all blocks were used for all the settings.} Posts within the same setting were also randomised, and shown to moderators in batches of 20 examples, one per page, on an internal annotation platform. 

Moderators did not undertake any training for this task. We asked them to judge whether a post violated the policy, underlining not to judge whether the explanation was correct. We also informed them that annotation times were being recorded. Finally, we provided moderators with one example for each scenario, to illustrate what the annotation interface would look like. We ran a pilot study with one moderator to assess the clarity of the interface and the soundness of our mapping of PLEAD annotations onto internal policy rules (Appendix \ref{sec:policy-adapt}). Details of the pilot can be found in Appendix \ref{sec:pilot}.

\subsection{Evaluation Metrics}
The annotation platform allowed us to record the timestamps at which posts were shown to moderators and when they moved to the next post, so for each post we stored the number of seconds it took to express a judgment. 
We also report moderator accuracy 
but do not expect an improvement from showing explanations, since these are professional moderators 
with a high degree of accuracy. Note also the limitation in accuracy measurements as this involves 
comparing the decisions of professional moderators -- who are regarded by online social platforms to be the ground truth -- against crowdsourced annotations.


\section{Do Explanations Help Moderators?}
%
Before analysing speed, we discarded the first 20 instances (0.025\%) from each setting. We did this to provide a buffer to the moderators to adapt to a new setting and corresponding interface. Additionally we discarded for each moderator all data points with annotation time more than three standard deviations away from the moderator mean\footnote{The number of outliers was comparable across settings.}. When moderators were prompted only with the post, the fastest and slowest moderators achieved a mean annotation speed of, respectively, 6.58s/post and 45.03s/post.
To study the effect of generic and structured explanations on annotation time (\textit{time}) while taking into account individual differences we fitted two linear mixed effects models to the data from \emph{post-only} and \emph{post+policy} or \emph{post+tags}, respectively. We defined the two models as follows:
\begin{equation*}
   \text{time} \sim{} \text{length} + (1|\text{moderator})
\end{equation*}
\begin{equation*}
   \text{time} \sim{} \text{setting} + \text{length} + (1|\text{moderator})
\end{equation*}
where \textit{length} is the length of the post, \textit{setting} indicates whether the moderator was provided an explanation or not, and $(1|\text{moderator})$ accounts for individual differences of the moderators. We tested whether the explanations have a significant effect by testing whether the difference between the likelihood of these two models is significant using ANOVA. We found that in setting \emph{post+policy} explanations did not affect the annotation time: the estimated effect is $0.02 \pm 0.32$ s, and is not significant ($\chi^2(1)=0.005$, $p=.94$).
When using structured explanations (\emph{post+tags}) the estimated effect is $-1.34 \pm 0.32$ s and is highly significant ($\chi^2(1)=17.808$, $p<.001$), showing that moderators are faster with appropriate explanations.
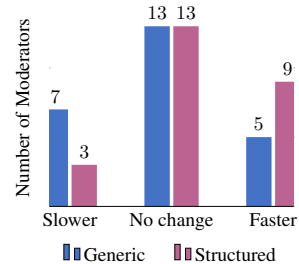
\begin{figure}
    \centering
    \resizebox{0.25\textwidth}{!}{
    \begin{tikzpicture}
  \centering
  \begin{axis}[
        ybar, axis on top,
        width=0.4\textwidth,
        bar width=0.5cm,
        enlarge y limits={value=.1,upper},
        ymin=0, ymax=13,
        yticklabel=\empty,
        axis x line*=bottom,
        y axis line style={opacity=0},
        tickwidth=0pt,
        enlarge x limits=true,
        legend style={
            at={(0.5,-0.17)},
            anchor=north,
            legend columns=-1,
            draw=none,
            /tikz/every even column/.append style={column sep=0.5cm}
        },
        ylabel={Number of Moderators},
        symbolic x coords={
           Slower,No change,Faster},
       xtick=data,
       nodes near coords={
        \pgfmathprintnumber[precision=0]{\pgfplotspointmeta}
       }
    ]
    \addplot [draw=none, fill=bar1] coordinates {
      (Slower,7)
      (No change, 13) 
      (Faster,5)};
   \addplot [draw=none,fill=bar2] coordinates {
      (Slower,3)
      (No change, 13) 
      (Faster,9)};

    \legend{Generic,Structured}
  \end{axis}
  \end{tikzpicture}
  }
  \caption{Effect of generic and structured explanations on the speed of each moderator (\textit{No change}: $|z| < 2$).}
    \label{tab:speed}
\end{figure}

We used a z-test to compare individual performances across the settings (Figure \ref{tab:speed}). When shown generic explanations 52\% of the moderators registered no significant change in speed (w.r.t. setting 1), 28\% had a significant loss in performance, and only 20\% improved. With structured explanations instead, 36\% of the moderators had a significant improvement, 52\% of the moderators registered no significant change, and 12\% performed worse than without explanations\footnote{One of these three moderators declared in the follow-on survey to have ignored the explanations.}.
We examined whether the different impact that explanations had on moderators was due to the experimental design by testing for correlations between said impact and the order in which the settings were shown to the moderators. With structured explanations, \textit{all} moderators who registered a loss in performance were shown this setting first and the Pearson correlation between the impact (represented as -1 for loss, 0 for no change, and 1 for improvement) and the round in which setting 3 was shown is .66 ($p<.001$). However, the same trend was not observed for generic explanations. Moderators who registered a loss in performance were shown \emph{post+policy} as either first or \textit{last}, and the correlation score is .41 ($p=.04$) (Appendix \ref{sec:setting-order}). We hypothesise that the posts from PLEAD might have been very different in language and topics from the ones moderators usually review, and therefore annotations in the first batch required moderators some extra adjustment time (regardless of the setting). However, the different trends observed for \emph{post+policy} and \emph{post+tags} demonstrate that the improvement recorded with structured explanations is not only related to the experimental design. Moreover, 
\emph{post+tags} is the setting that was shown as first 1 time more than the other settings (9 instead of 8), and 2 of the corresponding 9 moderators still registered a significant improvement.

We did not observe any correlation between the impact of explanations and the specific sample of 800 posts that was selected for each setting (-.06 for setting 2 and .09 for setting 3) (Appendix \ref{sec:sample-effect}).

Finally, we looked at accuracy to ensure that faster annotation did not come at the price of more mistakes. In \emph{post-only}, the highest and lowest recorded accuracy scores were 92.13\% and 73.13\%. We compared the accuracy of moderators across scenarios with a z-test between the accuracy of all moderators in setting 1 and 2 or 3. For both generic and structured explanations we did not observe a significant change ($z < 2$), not even when measuring accuracy only on not-hateful posts or hateful posts with wrong explanations (Appendix \ref{sec:accuracy}).

\section{Do Moderators Want Explanations?}
After the experiment was over, we asked the 25 moderators to complete a brief survey. A strong preference was expressed for the setting with structured explanations (84\%), while 8\% had no preference and 8\% preferred generic explanations (Appendix \ref{sec:survey}). When prompted with generic explanations, only 8\% of the moderators consistently took them into account, while 80\% only looked at the explanations when in doubt and 12\% ignored them. The picture changes for structured explanations, where 60\% of the moderators used them consistently, 32\% looked at them when in doubt, and 8\% ignored them. 48\% of the moderators declared that the posts shown in this study were different from the ones they usually moderate. They differed in the use of abbreviations, slang and jargon, but also in topics, as the policy covers many phenomena and hate speech is not the most frequent. This supports our hypothesis that moderators required some extra adjustment time in the first setting. 

\section{Conclusions}
In this work we investigated the impact of explainable NLP models on the decision speed of social media moderators.
Our experiments showed that explanations make moderators faster, but only when presented in the appropriate format. Generic explanations have no impact on decision time and are likely to be ignored, while structured explanations 
made moderators faster by 1.34 s/instance. 
A follow-on survey further revealed that moderators prefer structured explanations over generic or none.
These results were obtained simulating a model accuracy of 80\%, with 10\% of the posts misclassified as policy violations, and 10\% correctly classified but associated with wrong explanations. Such accuracy is beyond the capabilities of available models, and yet resulted in criticism from the moderators who spotted the inaccuracies. We hope this study can encourage researchers to improve abuse detection models that produce structured explanations.

\section{Limitations}
In this work we focused on hate speech, but there may be other content forbidden by a platform's terms that this work did not test. We focused on textual content and limited the study to English posts. These choices were merely driven by the lack of explainable multimodal and multilingual datasets for the task of integrity, or hate speech detection. Restricting the scope to English hate speech allowed us to compare the effects of different types of explanations on the same posts. We hope that the results reported in this study can promote the collection of structured explanations for new and existing multimodal or multilingual datasets.

\section{Ethical Considerations}
All the annotations in this study were produced by content moderators regularly employed at an online social platform. Although the posts they were asked to judge came from a public dataset and are different in style from the ones they usually review, dealing with hate speech is part of their role and they have been trained for handling such content.
No user data from said platform was used in this study, and all annotations of the public posts have been released in anonymised format\footnote{\cameraready{\url{https://github.com/Ago3/structured_explanations_make_moderators_faster}}} to protect the identity of the moderators.
We did not collect personal information about the moderators to protect their privacy, \cameraready{as 1) we are analyzing hate speech in a prescriptive paradigm that assumes the existence of a single ground truth and therefore it makes it less relevant to consider the demographics of individual annotators; 2) it would require asking platform employees for their protected characteristics. 
}

\section*{Acknowledgements}
We would like to thank Maryna Diakonova and the 25 Snapchat moderators who participated in our study.
This work was supported in part by Huawei and the UKRI Centre for Doctoral Training in Natural Language Processing, funded by the UKRI (grant EP/S022481/1) and the University of Edinburgh, School of Informatics. Lapata gratefully acknowledges
the support of the UK Engineering and Physical Sciences Research
Council (grant EP/W002876/1) 
and the European Research Council (award 681760).

\begin{figure}[h!]
    \centering
    \begin{subfigure}[b]{0.15\textwidth}
        \centering
        \includegraphics[height=6.2mm]{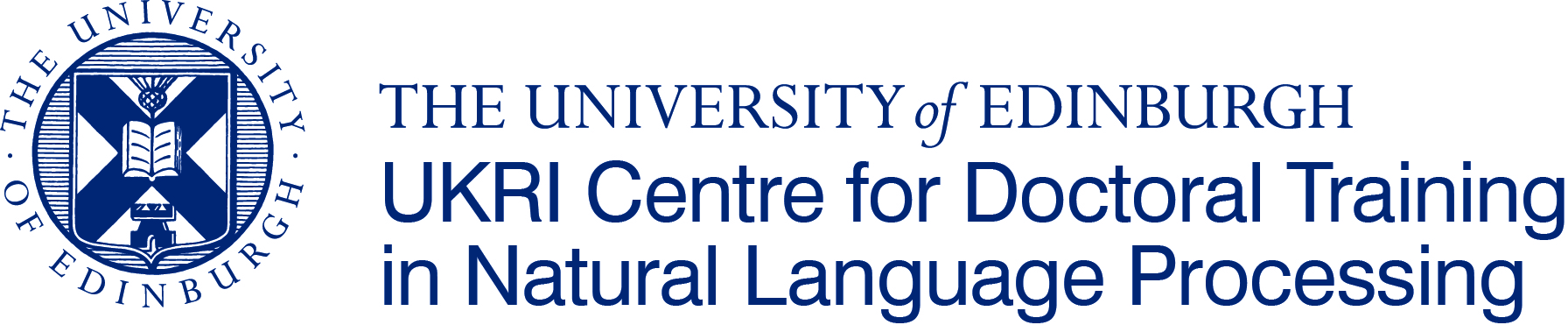}
    \end{subfigure}
    \hfill
    \begin{subfigure}[b]{0.08\textwidth}
        \centering
        \includegraphics[height=5.7mm]{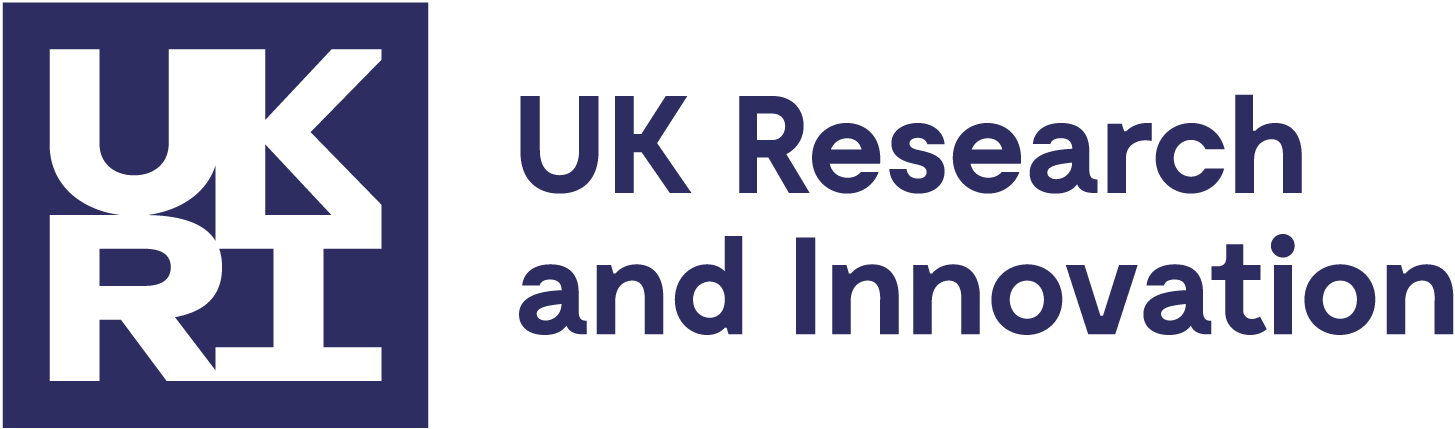}
    \end{subfigure}
    \hfill
    \begin{subfigure}[b]{0.15\textwidth}
        \centering
        \includegraphics[height=6.2mm]{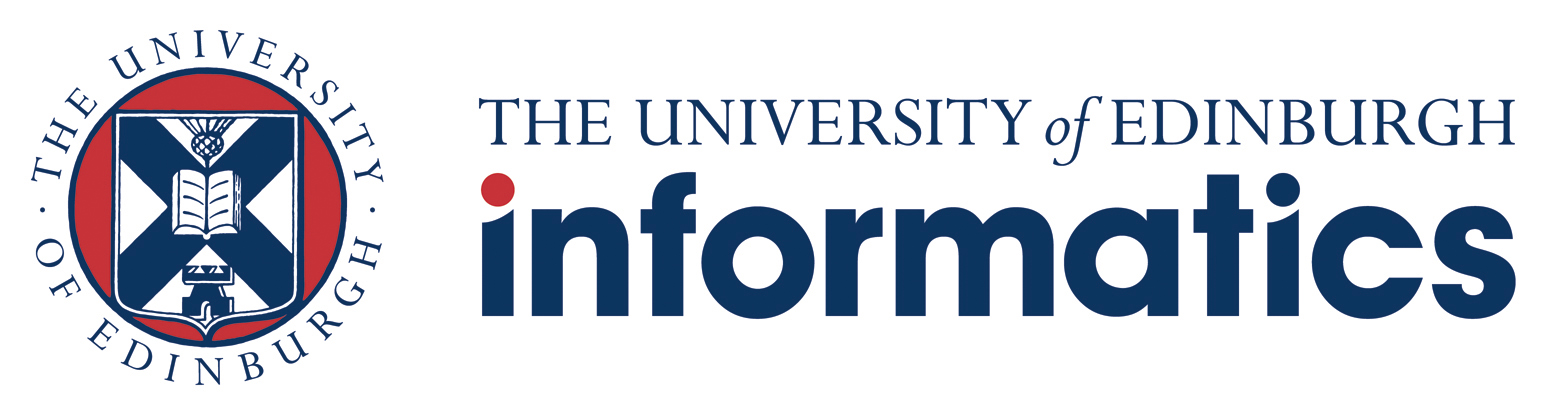}
    \end{subfigure}
\end{figure}


\bibliography{custom}

\appendix
\section{Experimental Design}
\label{sec:exp-design}

\subsection{PLEAD}
\label{sec:plead}
PLEAD is an extension of the LFTW dataset \cite{2021-vidgen-lftw} where the hateful and not-hateful posts have been enriched with span-level annotations for the task of intent classification and slot filling. Slots represent properties like ``target'' and ``protected characteristic'', while intents are policy rules
or guidelines (e.g., ``dehumanisation'').
PLEAD contains 3,535 posts, 25\% of which are not-hateful, while the remaining posts correspond to the intents of dehumanisation (25\%), threatening (17\%), derogation (28\%) and support of hate crimes (5\%).

\subsection{Policy Adaptation}
\label{sec:policy-adapt}
PLEAD was annotated using the codebook for hate speech annotations
designed by the Alan Turing Institute \cite{2021-vidgen-lftw}, and although everything that is labelled as hate speech in PLEAD also violates social media policies\footnote{e.g., \url{https://transparency.fb.com/policies/community-standards/} or \url{https://values.snap.com/en-GB/privacy/transparency/community-guidelines}}, the converse does not apply. Specifically, threats and harassment are not allowed by social media even when targeted at groups that are not protected. Therefore we manually reviewed all the not-hateful posts containing threats or derogatory expressions in the parse tree and labelled as policy violations all the posts in which such expressions are targeted at people.
For the second setting, where posts are shown together with a description of the violated rule, we adapted the wording in the explanations to match the internal policy the moderators are familiar with.

\subsection{Error Simulation}
\label{sec:error-simulation}
\begin{table*}[th]
\centering
\resizebox{\textwidth}{!}{%
\begin{tabular}{ll}
\toprule
\multicolumn{1}{c}{\bf Original Explanation}  & \multicolumn{1}{c}{\bf Wrong Explanation}\\
\midrule
\multirow{4}{*}{\shortstack[l]{My agenda is buying a new car and driving everywhere I can to see the world\\with my trans boyfriend apart from $[$\texttt{ProtectedCharacteristic}\hspace{.2cm}
\textit{muslim}$]$\\countries, they are full of $[$\texttt{Target, DehumanisingComparison}\hspace{.2cm} \textit{terrorists}$]$.\\And we just bought one yesterday so be careful! Agenda in motion!}} &
\multirow{4}{*}{\shortstack[l]{My agenda is buying a new car and driving everywhere I can to see the world\\
 with $[$\texttt{Target, ProtectedCharacteristic}\hspace{.2cm} \textit{my trans boyfriend}$]$ apart from muslim\\countries, they are full of $[$\texttt{DehumanisingComparison}\hspace{.2cm} \textit{terrorists}$]$.\\And we just bought one yesterday so be careful! Agenda in motion!}}\\
\multicolumn{2}{c}{} \\
& \\
& \\
\midrule
\multirow{2}{*}{\shortstack[l]{$[$\texttt{Target, ProtectedCharacteristic}\hspace{.2cm}
\textit{Women}$]$ like me\\$[$\texttt{DehumanisingComparison}\hspace{.2cm} \textit{are scum}$]$? $[$\texttt{NegativeStance}\hspace{.2cm} \textit{Shut up}$]$}} &
\multirow{2}{*}{\shortstack[l]{$[$\texttt{Target, ProtectedCharacteristic}\hspace{.2cm}
\textit{Women}$]$ like me\\$[$\texttt{DehumanisingComparison}\hspace{.2cm} \textit{are scum}$]$? Shut up}}\\
\multicolumn{2}{c}{} \\
\bottomrule
\end{tabular}%
}
\caption{
Example of hateful post (first row) and not-hateful post (second row) with the original (left column) and generated wrong (right column) explanation.
\label{tab:example-wrong-explanations}}
\end{table*}
To simulate model errors we tweaked some of the parse trees from PLEAD. Not-hateful posts are labelled as such when they lack at least one tag in the parse tree to violate the policy (e.g., they do not contain a reference to a protected group) or when a span of text tagged as negative stance is present (e.g. they quote a hateful expression only to disagree with it). For the 10\% of the posts that we sampled among the not-hateful ones, we either hallucinated new tagged spans, or deleted a negative stance tag. To prevent the moderators from associating obviously inaccurate explanations with the not-hateful class, we also simulated mistakes in the explanations of 10\% of the hateful posts. For these instances we dropped one tagged span from the parse tree, and hallucinated a new one to keep a policy violation. We first used heuristics to generate better explanations by only selecting noun phrases when hallucinating tags like \textit{target} and verb phrases for, e.g., \textit{threat}. We then manually reviewed and edited the modified explanations. Examples of wrong explanations are shown in Table~\ref{tab:example-wrong-explanations}.

\subsection{Pilot Study}
\label{sec:pilot}
We ran a pilot study with one of the moderators to assess the clarity of the interface and the soundness of our mapping of PLEAD annotations onto internal policy rules. We intentionally decided against asking more of the moderators to take the pilot, to avoid learning effects that could affect the final results. The pilot moderator was shown the same 100 posts in each setting, and achieved an accuracy of 93\% in all of them. This suggests that the interface did not confuse the moderator into judging the coherence of the explanations instead of the posts themselves, and that the mapping between the policies was accurate.
Since the posts were the same, it is not meaningful to compare the speed across the settings. The moderator started from setting 3 (posts + structured explanations) and took on average 13.11 seconds per instance. The re-annotation of the same posts in the following settings was faster, as expected.

\section{Effect of Settings Order}
\label{sec:setting-order}
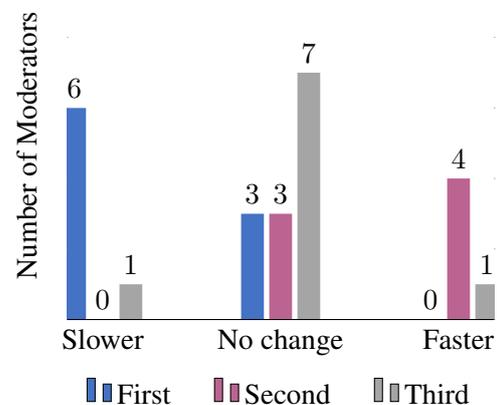
\begin{figure}[!h]
    \centering
    \begin{tikzpicture}
  \centering
  \begin{axis}[
        ybar, axis on top,
        width=0.45\textwidth,
        bar width=0.3cm,
        enlarge y limits={value=.1,upper},
        ymin=0, ymax=9,
        yticklabel=\empty,
        axis x line*=bottom,
        y axis line style={opacity=0},
        tickwidth=0pt,
        enlarge x limits=true,
        legend style={
            at={(0.5,-0.15)},
            anchor=north,
            legend columns=-1,
            draw=none,
            /tikz/every even column/.append style={column sep=0.5cm}
        },
        ylabel={Number of Moderators},
        symbolic x coords={
           Slower,No change,Faster},
       xtick=data,
       nodes near coords={
        \pgfmathprintnumber[precision=0]{\pgfplotspointmeta}
       }
    ]
    \addplot [draw=none, fill=bar1] coordinates {
      (Slower,6)
      (No change, 3) 
      (Faster,0)};
  \addplot [draw=none,fill=bar2] coordinates {
      (Slower,0)
      (No change, 3) 
      (Faster,4)};
   \addplot [draw=none,fill=bar3] coordinates {
      (Slower,1)
      (No change, 7) 
      (Faster,1)};

    \legend{First,Second,Third}
  \end{axis}
  \end{tikzpicture}
  \caption{Effect of generic explanations on the speed of individual moderators, grouped depending on which round they were shown this setting (\textit{No change}: $|z| < 2$).}
    \label{tab:generdic-order}
\end{figure}
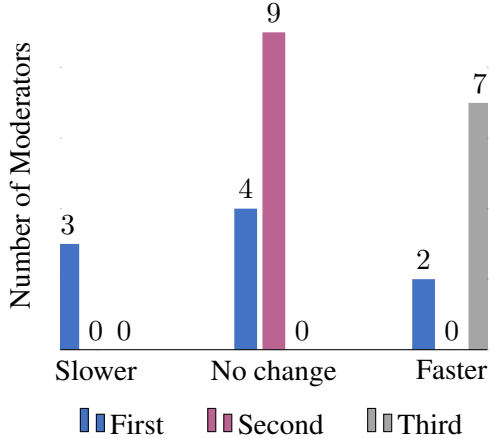
\begin{figure}[h]
    \centering
    \begin{tikzpicture}
  \centering
  \begin{axis}[
        ybar, axis on top,
        width=0.45\textwidth,
        bar width=0.3cm,
        enlarge y limits={value=.1,upper},
        ymin=0, ymax=9,
        yticklabel=\empty,
        axis x line*=bottom,
        y axis line style={opacity=0},
        tickwidth=0pt,
        enlarge x limits=true,
        legend style={
            at={(0.5,-0.15)},
            anchor=north,
            legend columns=-1,
            draw=none,
            /tikz/every even column/.append style={column sep=0.5cm}
        },
        ylabel={Number of Moderators},
        symbolic x coords={
           Slower,No change,Faster},
       xtick=data,
       nodes near coords={
        \pgfmathprintnumber[precision=0]{\pgfplotspointmeta}
       }
    ]
    \addplot [draw=none, fill=bar1] coordinates {
      (Slower,3)
      (No change, 4) 
      (Faster,2)};
  \addplot [draw=none,fill=bar2] coordinates {
      (Slower,0)
      (No change, 9) 
      (Faster,0)};
   \addplot [draw=none,fill=bar3] coordinates {
      (Slower,0)
      (No change, 0) 
      (Faster,7)};

    \legend{First,Second,Third}
  \end{axis}
  \end{tikzpicture}
  \caption{Effect of structured explanations on the speed of individual moderators, grouped depending on which round they were shown this setting (\textit{No change}: $|z| < 2$).}
    \label{tab:structured-order}
\end{figure}

We tested for correlations between the impact that explanations had on moderators speed and the order in which the settings were shown to the moderators. Figure~\ref{tab:structured-order} shows that with structured explanations, \textit{all} moderators who registered a loss in performance were shown this setting first. However, the same trend was not observed for generic explanations, where moderators who registered a loss in performance were shown \emph{post+policy} as either first or \textit{last} (Figure~\ref{tab:generdic-order}).

\section{Effect of Post Samples}
\label{sec:sample-effect}
\begin{figure}[h]
    \centering
    \begin{tikzpicture}
  \centering
  \begin{axis}[
        ybar, axis on top,
        width=0.45\textwidth,
        bar width=0.3cm,
        enlarge y limits={value=.1,upper},
        ymin=0, ymax=5,
        yticklabel=\empty,
        axis x line*=bottom,
        y axis line style={opacity=0},
        tickwidth=0pt,
        enlarge x limits=true,
        legend style={
            at={(0.5,-0.15)},
            anchor=north,
            legend columns=-1,
            draw=none,
            /tikz/every even column/.append style={column sep=0.5cm}
        },
        ylabel={Number of Moderators},
        symbolic x coords={
           Slower,No change,Faster},
       xtick=data,
       nodes near coords={
        \pgfmathprintnumber[precision=0]{\pgfplotspointmeta}
       }
    ]
    \addplot [draw=none, fill=bar1] coordinates {
      (Slower,2)
      (No change, 4) 
      (Faster,2)};
  \addplot [draw=none,fill=bar2] coordinates {
      (Slower,2)
      (No change, 5) 
      (Faster,1)};
   \addplot [draw=none,fill=bar3] coordinates {
      (Slower,3)
      (No change, 4) 
      (Faster,2)};

    \legend{Sample 1,Sample 2,Sample 3}
  \end{axis}
  \end{tikzpicture}
  \caption{Effect of generic explanations on the speed of individual moderators, grouped depending on which sample of 800 posts was used for this setting (\textit{No change}: $|z| < 2$).}
    \label{tab:generdic-sample}
\end{figure}
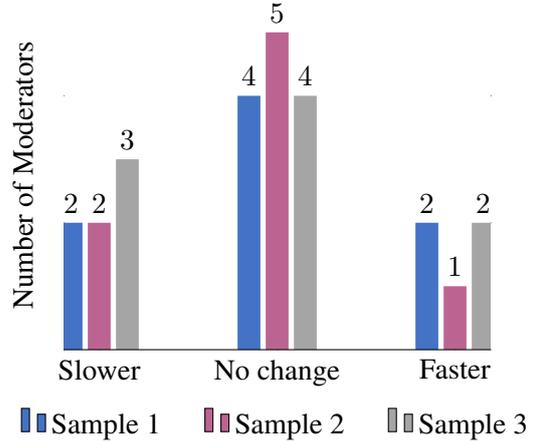
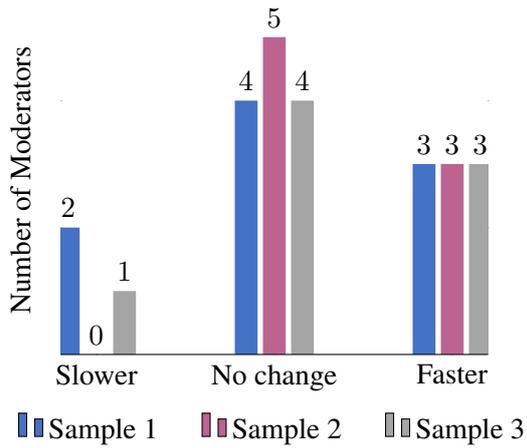
\begin{figure}[h]
    \centering
    \begin{tikzpicture}
  \centering
  \begin{axis}[
        ybar, axis on top,
        width=0.45\textwidth,
        bar width=0.3cm,
        enlarge y limits={value=.1,upper},
        ymin=0, ymax=5,
        yticklabel=\empty,
        axis x line*=bottom,
        y axis line style={opacity=0},
        tickwidth=0pt,
        enlarge x limits=true,
        legend style={
            at={(0.5,-0.15)},
            anchor=north,
            legend columns=-1,
            draw=none,
            /tikz/every even column/.append style={column sep=0.5cm}
        },
        ylabel={Number of Moderators},
        symbolic x coords={
           Slower,No change,Faster},
       xtick=data,
       nodes near coords={
        \pgfmathprintnumber[precision=0]{\pgfplotspointmeta}
       }
    ]
    \addplot [draw=none, fill=bar1] coordinates {
      (Slower,2)
      (No change, 4) 
      (Faster,3)};
  \addplot [draw=none,fill=bar2] coordinates {
      (Slower,0)
      (No change, 5) 
      (Faster,3)};
   \addplot [draw=none,fill=bar3] coordinates {
      (Slower,1)
      (No change, 4) 
      (Faster,3)};

    \legend{Sample 1,Sample 2,Sample 3}
  \end{axis}
  \end{tikzpicture}
  \caption{Effect of structured explanations on the speed of individual moderators, grouped depending on which sample of 800 posts was used for this setting (\textit{No change}: $|z| < 2$).}
    \label{tab:structured-sample}
\end{figure}

We tested for correlations between the impact that explanations had on moderators speed and the specific sample of 800 posts that was selected for each setting. As Figure~\ref{tab:generdic-sample} and \ref{tab:structured-sample} show no clear pattern emerged, and the correlation between impact and sample was -.06 for \emph{post+label} and .09 for \emph{post+tags}.

\section{Accuracy}
\label{sec:accuracy}

\begin{figure}[h]
    \centering
    \begin{tikzpicture}
    \begin{axis}[%
    ymin=0, ymax=100,
    width=0.48\textwidth,
    ylabel={Accuracy (\%)},
    xlabel={Moderator ID},
    legend style={
            at={(0.5,-0.2)},
            anchor=north,
            legend columns=-1,
            draw=none,
            /tikz/every even column/.append style={column sep=0.5cm}
        },
    scatter/classes={%
        a={mark=*,bar1},
        b={mark=*,bar2},
        c={mark=*,bar3}}]
    \addplot[scatter,only marks,%
        scatter src=explicit symbolic]%
    table[meta=style] {
    x y label style
    0 92.125 None a
    1 89.5 None a
    2 91. None a
    3 88. None a
    4 90.875 None a
    5 91.125 None a
    6 88.5 None a
    7 78.5 None a
    8 91.25 None a
    9 90.875 None a
    10 86.25 None a
    11 87.125 None a
    12 73.125 None a
    13 83.375 None a
    14 91.375 None a
    15 89.125 None a
    16 75.25 None a
    17 84.25 None a
    18 91.5 None a
    19 88.5 None a
    20 91.5 None a
    21 90.625 None a
    22 80.75 None a
    23 91.875 None a
    24 83.875 None a
        };
    \addplot[scatter,only marks,%
        scatter src=explicit symbolic]%
    table[meta=style] {
    x y label style
    0 91.625 Generic b
    1 87.875 Generic b
    2 91.875 Generic b
    3 90.25 Generic b
    4 90.25 Generic b
    5 91.5 Generic b
    6 91.25 Generic b
    7 85.625 Generic b
    8 92.125 Generic b
    9 90.625 Generic b
    10 81. Generic b
    11 87.125 Generic b
    12 73.375 Generic b
    13 89.5 Generic b
    14 91. Generic b
    15 91.625 Generic b
    16 87.125 Generic b
    17 87. Generic b
    18 92. Generic b
    19 88.25 Generic b
    20 89.5 Generic b
    21 90 Generic b
    22 85. Generic b
    23 89.25 Generic b
    24 91.375 Generic b
        };
    \addplot[scatter,only marks,%
        scatter src=explicit symbolic]%
    table[meta=style] {
    x y label style
    0 92.5 Structured c
    1 90.375 Structured c
    2 91.375 Structured c
    3 86. Structured c
    4 89.25 Structured c
    5 89.25 Structured c
    6 93.5 Structured c
    7 88.5 Structured c
    8 91. Structured c
    9 91.875 Structured c
    10 81.375 Structured c
    11 90.625 Structured c
    12 78.75 Structured c
    13 81.375 Structured c
    14 88. Structured c
    15 93.25 Structured c
    16 86.125 Structured c
    17 85.25 Structured c
    18 91.75 Structured c
    19 89.875 Structured c
    20 92.5 Structured c
    21 90.5 Structured c
    22 81.75 Structured c
    23 83.625 Structured c
    24 92.75 Structured c
        };
        \legend{None,Generic,Structured}
    \end{axis}
    \end{tikzpicture}
    \caption{Accuracy score achieved by each moderator with no, generic or structured explanations on the 3 different samples of 800 posts.}
    \label{fig:accuracy}
\end{figure}
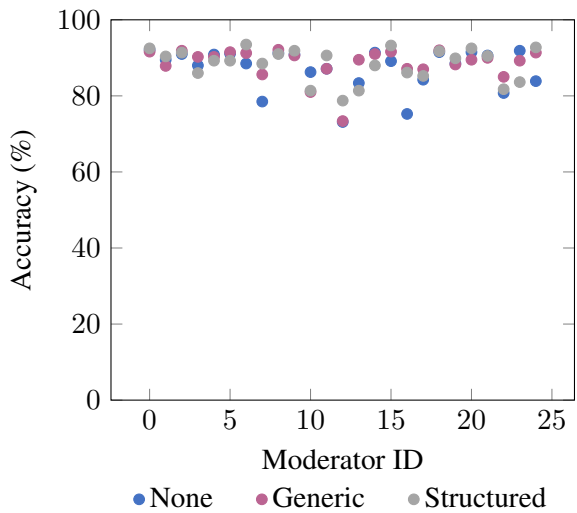

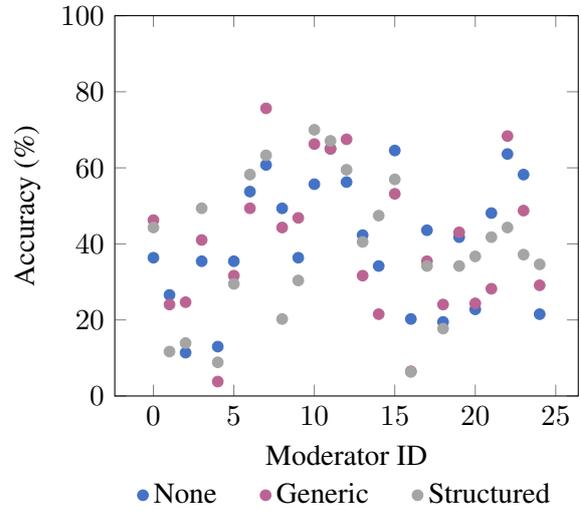
\begin{figure}[h]
    \centering
    \begin{tikzpicture}
    \begin{axis}[%
    ymin=0, ymax=100,
    width=0.48\textwidth,
    ylabel={Accuracy (\%)},
    xlabel={Moderator ID},
    legend style={
            at={(0.5,-0.2)},
            anchor=north,
            legend columns=-1,
            draw=none,
            /tikz/every even column/.append style={column sep=0.5cm}
        },
    scatter/classes={%
        a={mark=*,bar1},
        b={mark=*,bar2},
        c={mark=*,bar3}}]
    \addplot[scatter,only marks,%
        scatter src=explicit symbolic]%
    table[meta=style] {
    x y label style
    0 36.363636363636365 None a
    1 26.582278481012656 None a
    2 11.392405063291139 None a
    3 35.443037974683544 None a
    4 12.987012987012986 None a
    5 35.443037974683544 None a
    6 53.75 None a
    7 60.75949367088608 None a
    8 49.35064935064935 None a
    9 36.363636363636365 None a
    10 55.69620253164557 None a
    11 65. None a
    12 56.25 None a
    13 42.30769230769231 None a
    14 34.177215189873417 None a
    15 64.55696202531646 None a
    16 20.253164556962025 None a
    17 43.58974358974359 None a
    18 19.480519480519481 None a
    19 41.77215189873418 None a
    20 22.784810126582278 None a
    21 48.10126582278481 None a
    22 63.63636363636364 None a
    23 58.22784810126582 None a
    24 21.518987341772153 None a
        };
    \addplot[scatter,only marks,%
        scatter src=explicit symbolic]%
    table[meta=style] {
    x y label style
    0 46.25 Generic b
    1 24.050632911392406 Generic b
    2 24.675324675324675 Generic b
    3 41.025641025641024 Generic b
    4 03.79746835443038 Generic b
    5 31.645569620253167 Generic b
    6 49.36708860759494 Generic b
    7 75.64102564102564 Generic b
    8 44.30379746835443 Generic b
    9 46.835443037974683 Generic b
    10 66.25 Generic b
    11 65. Generic b
    12 67.5 Generic b
    13 31.645569620253167 Generic b
    14 21.518987341772153 Generic b
    15 53.16455696202531 Generic b
    16 06.493506493506493 Generic b
    17 35.443037974683544 Generic b
    18 24.050632911392406 Generic b
    19 43.037974683544306 Generic b
    20 24.358974358974358 Generic b
    21 28.205128205128205 Generic b
    22 68.35443037974683 Generic b
    23 48.75 Generic b
    24 29.11392405063291 Generic b
        };
    \addplot[scatter,only marks,%
        scatter src=explicit symbolic]%
    table[meta=style] {
    x y label style
    0 44.30379746835443 Structured c
    1 11.688311688311688 Structured c
    2 13.924050632911392 Structured c
    3 49.36708860759494 Structured c
    4 08.860759493670886 Structured c
    5 29.48717948717949 Structured c
    6 58.22784810126582 Structured c
    7 63.29113924050633 Structured c
    8 20.253164556962025 Structured c
    9 30.37974683544304 Structured c
    10 70. Structured c
    11 67.08860759493671 Structured c
    12 59.49367088607594 Structured c
    13 40.50632911392405 Structured c
    14 47.435897435897434 Structured c
    15 56.9620253164557 Structured c
    16 06.329113924050633 Structured c
    17 34.177215189873417 Structured c
    18 17.721518987341772 Structured c
    19 34.177215189873417 Structured c
    20 36.70886075949367 Structured c
    21 41.77215189873418 Structured c
    22 44.30379746835443 Structured c
    23 37.17948717948718 Structured c
    24 34.615384615384615 Structured c
        };
        \legend{None,Generic,Structured}
    \end{axis}
    \end{tikzpicture}
    \caption{Accuracy score achieved by each moderator with no, generic or structured explanations on the 80 not-hateful instances of the 3 different samples.}
    \label{fig:accuracy-nh}
\end{figure}

\begin{figure}[h]
    \centering
    \begin{tikzpicture}
    \begin{axis}[%
    ymin=0, ymax=100,
    width=0.48\textwidth,
    ylabel={Accuracy (\%)},
    xlabel={Moderator ID},
    legend style={
            at={(0.5,-0.2)},
            anchor=north,
            legend columns=-1,
            draw=none,
            /tikz/every even column/.append style={column sep=0.5cm}
        },
    scatter/classes={%
        a={mark=*,bar1},
        b={mark=*,bar2},
        c={mark=*,bar3}}]
    \addplot[scatter,only marks,%
        scatter src=explicit symbolic]%
    table[meta=style] {
    x y label style
    0 97.5 None a
    1 97.5 None a
    2 100 None a
    3 96.25 None a
    4 98.75 None a
    5 96.25 None a
    6 90. None a
    7 80. None a
    8 96.25 None a
    9 96.25 None a
    10 93.75 None a
    11 90. None a
    12 83.75 None a
    13 82.5 None a
    14 97.5 None a
    15 88.75 None a
    16 81.25 None a
    17 86.25 None a
    18 100 None a
    19 98.75 None a
    20 100 None a
    21 98.75 None a
    22 83.75 None a
    23 93.75 None a
    24 91.25 None a
        };
    \addplot[scatter,only marks,%
        scatter src=explicit symbolic]%
    table[meta=style] {
    x y label style
    0 96.25 Generic b
    1 95. Generic b
    2 98.75 Generic b
    3 92.5 Generic b
    4 100 Generic b
    5 97.5 Generic b
    6 97.5 Generic b
    7 91.25 Generic b
    8 96.25 Generic b
    9 93.75 Generic b
    10 88.75 Generic b
    11 87.5 Generic b
    12 76.25 Generic b
    13 93.75 Generic b
    14 98.75 Generic b
    15 97.5 Generic b
    16 93.75 Generic b
    17 92.5 Generic b
    18 98.75 Generic b
    19 97.5 Generic b
    20 98.75 Generic b
    21 97.5 Generic b
    22 91.25 Generic b
    23 95. Generic b
    24 98.75 Generic b
        };
    \addplot[scatter,only marks,%
        scatter src=explicit symbolic]%
    table[meta=style] {
    x y label style
    0 96.25 Structured c
    1 100 Structured c
    2 100 Structured c
    3 92.5 Structured c
    4 97.5 Structured c
    5 96.25 Structured c
    6 92.5 Structured c
    7 95. Structured c
    8 98.75 Structured c
    9 96.25 Structured c
    10 76.25 Structured c
    11 93.75 Structured c
    12 77.5 Structured c
    13 87.5 Structured c
    14 88.75 Structured c
    15 95. Structured c
    16 91.25 Structured c
    17 93.75 Structured c
    18 100 Structured c
    19 98.75 Structured c
    20 98.75 Structured c
    21 96.25 Structured c
    22 95. Structured c
    23 87.5 Structured c
    24 100 Structured c
        };
        \legend{None,Generic,Structured}
    \end{axis}
    \end{tikzpicture}
    \caption{Accuracy score achieved by each moderator with no, generic or structured explanations on the 80 hateful instances of the 3 different samples that were shown with wrong explanations.}
    \label{fig:accuracy-hateful-wrong}
\end{figure}
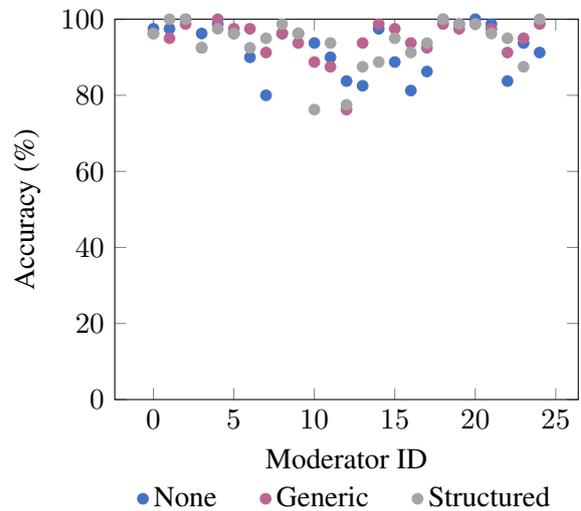

We compared the accuracy of moderators across scenarios with a z-test between the accuracy of all moderators in setting 1 (\emph{post-only}) and 2 (\emph{post+policy}) or 3 (\emph{post+label}). For both generic and structured explanations we did not observe a significant change ($z < 2$, Figure~\ref{fig:accuracy}), not even when measuring accuracy only on not-hateful posts (Figure~\ref{fig:accuracy-nh}) or hateful posts with wrong explanations (Figure~\ref{fig:accuracy-hateful-wrong}).

\section{Moderators' Preference}
\label{sec:survey}

\begin{figure}[h]
    \centering
    \resizebox{0.48\textwidth}{!}{
    \begin{tikzpicture}
    \pie{8/{No preference}, 8/Generic, 84/Structured}
    \end{tikzpicture}}
    \caption{We asked the 25 moderators whether they preferred the setting with generic explanations, structured explanations, or had no preference. The great majority preferred the setting with structured explanations.}
    \label{fig:preferred-setting}
\end{figure}
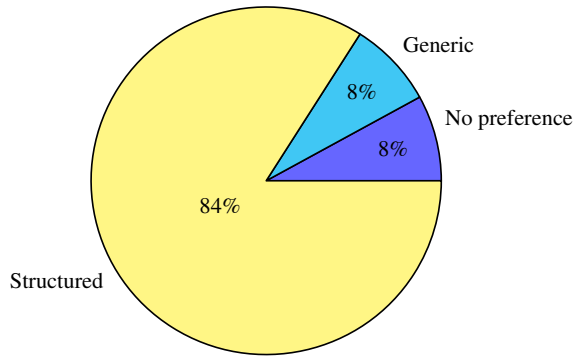

\begin{figure}[h]
    \centering
    \resizebox{0.48\textwidth}{!}{
    \begin{tikzpicture}
    \pie{12/No, 80/Only when in doubt, 8/Yes}
    \end{tikzpicture}}
    \caption{We asked the 25 moderators whether they used the generic explanations or ignored them. 80\% of the moderators declared to have used the explanations only when in doubt, and a further 12\% ignored the explanations.}
    \label{fig:generic-expl}
\end{figure}
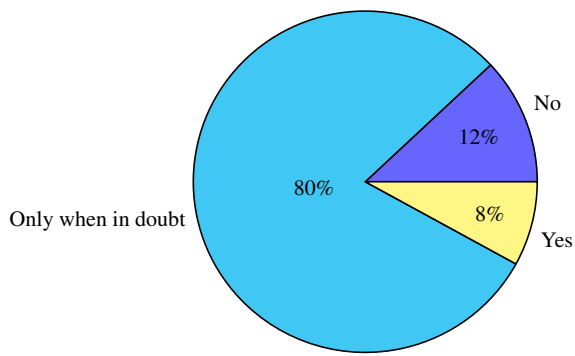

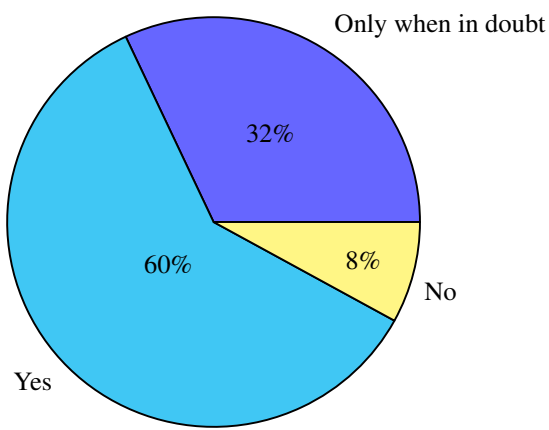
\begin{figure}[h]
    \centering
    \resizebox{0.48\textwidth}{!}{
    \begin{tikzpicture}
    \pie{32/Only when in doubt, 60/Yes, 8/No}
    \end{tikzpicture}}
    \caption{We asked the 25 moderators whether they used the structured explanations or ignored them. 60\% of the moderators declared to have used the explanations consistently, and a further 32\% relied on them when in doubt.}
    \label{fig:structured-expl}
\end{figure}

Figure \ref{fig:preferred-setting} summarises the moderators' preferences among the three settings. Only 8\% of the moderators expressed a preference for generic explanations, and this is coherent to the level of engagement that this type of explanations registered (Figure \ref{fig:generic-expl}). 84\% of the moderators expressed a preference for the structured explanations, with only 8\% who declared to have ignored the explanations during the annotation (Figure \ref{fig:structured-expl}). The criticisms raised about these explanations concerned their accuracy and the need to sometimes still read the whole post to grasp the context in which the highlighted expressions were used. Overall moderators did not think the design of the structured explanations could be further improved to optimise their decision speed. They stressed the importance of using the explanations as a guide while still reading the posts for context, leaving no margin for improvement on this metric.

When asked what the most common reasons were for them to be unsure about how to judge a post during their regular job, they indicated slang, unknown words/symbols and the lack of cultural context. Combining structured explanations with additional free-text explanations could be a way to support moderators when judging complex posts, improving their accuracy (but not speed).

\end{document}